\documentclass[10pt,journal,compsoc]{IEEEtran}
\usepackage{amsmath,amsfonts}
\usepackage{algpseudocode}
\usepackage{algorithm}
\usepackage{array}
\usepackage[caption=false,font=normalsize,labelfont=sf,textfont=sf]{subfig}
\usepackage{textcomp}
\usepackage{stfloats}
\usepackage{tabularx}
\usepackage{url}
\usepackage{verbatim}
\usepackage{graphicx}
\usepackage{cite}
\usepackage{amsmath}
\usepackage{amsthm}
\usepackage{booktabs}
\begin{document}

\title{Exploring New Frontiers in Agricultural NLP: Investigating the Potential of Large Language Models for Food Applications}
\author{Saed Rezayi, Zhengliang Liu, Zihao Wu, Chandra Dhakal, Bao Ge, Haixing Dai, Gengchen Mai,\\ Ninghao Liu, Chen Zhen, Tianming Liu, Sheng Li,~\IEEEmembership{Senior Member,~IEEE}
\IEEEcompsocitemizethanks{\IEEEcompsocthanksitem Saed Rezayi, Zhengliang Liu, Zihao Wu, Haixing Dai, Ninghao Liu, and Tianming Liu are with the School of Computing, University of Georgia, Athens, GA.
E-mail: \{saedr, zl18864, Zihao.Wu1, Haixing.Dai, ninghao.liu, tliu\}@uga.edu \protect

\IEEEcompsocthanksitem Chandra Dhakal and Chen Zhen are with the Department of Agricultural and Applied Economics, University of Georgia, Athens, GA.
E-mail: \{chandra.dhakal25, zchen\}@uga.edu \protect

\IEEEcompsocthanksitem Bao Ge is with Shaanxi Normal University, China. 
E-mail: bob\_ge@snnu.edu.cn \protect

\IEEEcompsocthanksitem Gengchen Mai is with the Department of Geography, University of Georgia, Athens, GA. Email: gengchen.mai25@uga.edu \protect

\IEEEcompsocthanksitem Sheng Li is with the School of Data Science, University of Virginia, Charlottesville, VA. E-mail: shengli@virginia.edu
}
}


\maketitle

\begin{abstract}

This paper explores new frontiers in agricultural natural language processing by investigating the effectiveness of using food-related text corpora for pretraining transformer-based language models. In particular, we focus on the task of semantic matching, which involves establishing mappings between food descriptions and nutrition data. To accomplish this, we fine-tune a pre-trained transformer-based language model, AgriBERT, on this task, utilizing an external source of knowledge, such as the FoodOn ontology. To advance the field of agricultural NLP, we propose two new avenues of exploration: (1) utilizing GPT-based models as a baseline and (2) leveraging ChatGPT as an external source of knowledge. ChatGPT has shown to be a strong baseline in many NLP tasks, and we believe it has the potential to improve our model in the task of semantic matching and enhance our model's understanding of food-related concepts and relationships. Additionally, we experiment with other applications, such as cuisine prediction based on food ingredients, and expand the scope of our research to include other NLP tasks beyond semantic matching. Overall, this paper provides promising avenues for future research in this field, with potential implications for improving the performance of agricultural NLP applications.
\end{abstract}

\begin{IEEEkeywords}
Natural Language Processing, Language Models, ChatGPT, Food Applications, Semantic Matching
\end{IEEEkeywords}

\section{Introduction}
The United States Department of Agriculture (USDA) maintains the Food and Nutrient Database for Dietary Studies (FNDDS), a repository of nutrient values for foods and beverages consumed within the United States\footnote{\url{https://www.ars.usda.gov/northeast-area/beltsville-md-bhnrc/beltsville-human-nutrition-research-center/food-surveys-research-group/docs/fndds/}}. Additionally, extensive food policy research leverages household and retail scanner data on grocery purchases, such as the Nielsen data available through the Kilts Center for Marketing\footnote{Researcher(s)' own analyses calculated (or derived) based in part on data from Nielsen Consumer LLC and marketing databases provided through the NielsenIQ Datasets at the Kilts Center for Marketing Data Center at The University of Chicago Booth School of Business. The conclusions drawn from the NielsenIQ data are those of the researcher(s), and do not reflect the views of NielsenIQ. NielsenIQ is not responsible for, had no role in, and was not involved in analyzing and preparing the results reported herein.}. The integration of these databases, i.e., bridging food description from retail scanner data with the nutritional information database, is of paramount importance. This linkage can shed light on the relationship between retail food purchase patterns and community health, revealing differences between the diets of low and higher-income households across the entire diet spectrum. Consequently, it has the potential to shape future funding policies aimed at facilitating healthier food choices for low-income households.

In this extended version of our previous paper~\cite{rezayi2022agribert}, we seek to advance and apply Natural Language Processing (NLP) techniques to facilitate a more robust linkage between these two databases. One common strategy to address such challenges is semantic matching, a task that identifies whether two or more elements share similar meanings. Bi-encoders, which encode two input strings in the embedding space and subsequently calculate their similarity in a supervised manner, are frequently used for this purpose. Word embedding techniques serve as excellent candidates for this task, given their extensive use and recent advancements in semantic matching~\cite{kenter2015short}. Particularly, the advent of contextual word embeddings, wherein each word receives a vector representation based on its context, has led to considerable progress in numerous NLP tasks, including semantic matching. However, while word embeddings have demonstrated utility in many tasks, they are not always optimal for more complex semantic matching tasks as they may struggle with subtleties in domain-specific language and may not effectively capture broader contextual information.

Transformer-based language models, e.g., BERT~\cite{devlin2019bert}, have been widely used in research and practice to study computational linguistics and they have shown superior performance in a variety of applications including text classification~\cite{jin2020bert,shi2023chatgraph}, question answering~\cite{yang2019end}, named entity recognition~\cite{mai2023opportunities}, and many more. However, these models may not always generalize across domains when applied with default objectives, i.e., pre-training on generic corpora like Wikipedia. To address this issue, previous work has attempted to incorporate domain-specific knowledge into the language model through different strategies. One of the prominent approaches in the biomedical domain is BioBERT~\cite{lee2020biobert}, a BERT-based language model pretrained on a large corpus of biomedical literature. Motivated by the impressive performance of BioBERT, we use a large corpus of agricultural literature to train a language model for agricultural applications from scratch. The trained model will be further fine-tuned by the downstream tasks.

Another method to incorporate domain knowledge into the language model is to use an external source of knowledge bases such as a knowledge graph (KG). Knowledge graphs \cite{auer2007dbpedia,vrandevcic2012wikidata,noy2019industry,janowicz2022know,mai2022symbolic,qi2023evkg} are densely interconnecting (Web-scale) directed multi-graphs containing data across various domains. It includes rich sources of information that are carefully curated around objects called entities and their relations. A basic building block of a KG is called a triple which consists of a subject, a predicate, and an object. Previous work has attempted to inject triples into the sentences~\cite{liu2020k} for language model pretraining. However, injecting triples can introduce noise to the sentences which will mislead the underlying text encoder. To address this issue, we propose to add $n$ entities from an external knowledge source (i.e., a knowledge graph) based on similarity that can be obtained by various methods such as entity linking. This not only enhances the semantic space but also confines the vocabulary within the domain. In our extended study, we provide empirical evidence showing how modifying $n$ can impact the performance of the downstream task both quantitatively and qualitatively.

In this paper, we map retail scanner data, referred to as Nielsen data, to USDA descriptions by utilizing semantic matching. This process is conceptualized as an ``answer selection'' problem. In our unique framing, we treat Nielsen product descriptions as questions and USDA descriptions as potential answers, seeking to link each Nielsen product with the most suitable USDA description. This is different from traditional answer selection, where each question has a unique, limited set of answers. Here, we have a vast, shared pool of USDA descriptions serving as potential answers. To support our answer selection system, we use a pre-trained language model, enriching both the Nielsen and USDA descriptions by incorporating external knowledge during the fine-tuning phase. In this extended version, we further enhance our approach by leveraging the power of GPT-based large language models (LLMs). We utilize these LLMs to augment the answer selection process, similar to the concept of entity linking. Additionally, we explore the utilization of GPT-based LLMs as an independent baseline for comparison. By incorporating these advancements, we aim to explore alternative approaches and methodologies to further investigate the mapping process of food descriptions in retail scanner data to the nutritional information database. By incorporating GPT-based large language models into our analysis, we seek to provide valuable insights and novel perspectives that can contribute to the understanding of this mapping task. We thus make the following key contributions in this paper:

\begin{itemize}
    \item We collect a large-scale corpus of agricultural literature with more than 300 million tokens. This domain corpus has been instrumental to fine-tune generic BERT into AgriBERT. 
    \item We propose a knowledge graph-guided approach to augment the dataset for the answer selection component. We inject related entities into the sentences before the fine-tuning step.
    \item We investigate the integration of GPT-based large language models to gain new insights in the mapping of food descriptions in retail scanner data to the nutritional information database.
    \item AgriBERT substantially outperforms existing language models on USDA datasets in the task of answer selection. We plan to release our datasets to the community upon publication.  
\end{itemize}

The rest of the paper is organized as follows: in the next section, we discuss related works in language modeling in specific domains. Next, in Section \ref{sec:method}, we describe our proposed approach to train a language model in the agricultural domain and discuss how we inject external knowledge in the fine-tuning step. In Section \ref{sec:exp}, we introduce different datasets including our corpus for training a language model, the external sources of knowledge, and finally the food dataset to evaluate our language model. We conclude our paper in Section \ref{sec:con}.

\label{sec:intro}

\section{Related Works}

\subsection{Pre-trained Language Models}
In NLP, Pre-trained language models learn from large text corpora and build representations beneficial for various downstream tasks. In recent years, there are two successive generations of language models. Earlier models, such as Skip-Gram ~\cite{mikolov2013distributed} and GloVe ~\cite{pennington2014glove}, primarily focus on learning word embeddings from statistical patterns, semantic similarities, and syntactic relationships at the word level. With this first group of language embedding methods, polysemous words are mapped to the same representation, regardless of word contexts. For example, the word "bear" in "I see a bear" and "Rising car sales bear witness to population increase in this area" will not be distinguishable in the vector space. 

A later group of models, however, recognizes the importance of textual contexts and aims to learn context-dependent representations at the sentence level or higher. For example, CoVe ~\cite{McCann2017LearnedIT} utilizes an LSTM model trained for machine translation to encode contextualized word vectors. Another popular model, Bidirectional Encoder Representations from Transformers (BERT) ~\cite{devlin2019bert} is based on bidirectional transformers and pre-trained with Masked Language Modeling (MLM) and Next Sentence Prediction (NSP) tasks, both ideal training tasks for learning effective contextual representations from unlabelled data. It delivers exceptional performance and can be easily fine-tuned for downstream tasks. BERT has enjoyed wide acceptance from the NLP community and practitioners from other domains~\cite{chalkidis2020legal,liu2023context,wang2017coupled}. In particular, domain experts can build domain-specific BERT models \cite{rezayi2022clinicalradiobert,liu2022survey,gu2021domain} that cater to specific environments and task scenarios \cite{liao2023mask,cai2022coarse,koroteev2021bert}.

\subsection{Large Language Models}
The NLP field has witnessed the very recent rise of large language models (LLM), boasting parameter counts exceeding 100 billion. GPT-3 \cite{brown2020language}, a large scale model with 175 billion parameters, introduced a new paradigm for NLP. Although LLMs are still based on the Transformer architecture, there is an enormous increase in the training data, parameter count and overall model dimensions. The increased scale of LLMs enable powerful zero-shot and few-shot learning capabilities \cite{brown2020language,liu2023summary} through in-context learning \cite{brown2020language} that does not involve gradient updates. In addition, LLMs demonstrate emergent abilities (e.g., reasoning, mathematical capabilities) \cite{wei2022emergent,zhong2023chatabl} that are not available in smaller pre-trained models such as BERT and its variants. For example, GPT-4 delivers expert level performance on a wide range of examinations and benchmarks including USMLE \cite{openai2023gpt,nori2023capabilities}, LSAT \cite{openai2023gpt} or highly specialized domain evaluations such as radiation oncology physics exams \cite{holmes2023evaluating}. 

Some prominent examples of LLMs include Bloom \cite{scao2022bloom}, OPT \cite{zhang2022opt}, LLAMA \cite{touvron2023llama}, ChatGPT \cite{liu2023summary}, GPT-4 \cite{openai2023gpt} and Palm 2 \cite{anil2023palm}. Indeed, the fame of ChatGPT has revolutionized and popularized NLP, leading to new research \cite{ma2023impressiongpt,wu2023exploring,dai2023chataug,wang2023chatcad} and applications \cite{liao2023differentiate,liu2023deid} with LLMs as foundational models \cite{zhao2023brain}. 

\begin{table*}[t]
    \centering
    \caption{An example of how we propose to extend the dataset.}
    \label{tab:example}
    \resizebox{.65\textwidth}{!}{%
    \begin{tabular}{@{}lll@{}}\toprule
        Product Description & USDA Description & Label \\ \midrule
        domino white sugar granulated 1lb & salsa, red, commercially-prepared & False \\
        domino white sugar granulated 1lb & cookie-crisp & False \\
        domino white sugar granulated 1lb & sugar, white, granulated or lump & True \\ \bottomrule
    \end{tabular}
    }

\end{table*}

\subsection{Domain-Specific Language Models}
BERT has become a fundamental building block for training task-specific models. It can be further extended with domain-specific pre-training to achieve additional gains over general-purpose language models. 

Prior work has shown that language models perform better when the source and target domains are highly relevant ~\cite{lee2020biobert,gu2021domain,wang2023prompt,lu2023agi,zhao2023brain,liu2023summary}. In other words, pre-training BERT models with in-domain corpora can significantly improve overall performance on a wide variety of downstream tasks ~\cite{gu2021domain}. 

There is also a correlation between a model’s performance and the extent of domain-specific training ~\cite{gu2021domain}. In particular, Gu et al. ~\cite{gu2021domain} note that training models from scratch (i.e., not importing pre-trained weights from the original BERT model ~\cite{devlin2019bert} or any other existing BERT-based models) is more effective than simply fine-tuning an existing BERT model with domain-specific data. 

In this paper, agricultural text such as food-related research papers is considered in-domain while other sources such as Wikipedia and news corpus are regarded as out-domain or general domain. Our primary approach is in line with training-from-scratch with in-domain data. 

\subsection{Augmenting Pre-trained Language Models}
Data augmentation refers to the practice of increasing training data size and diversity without collecting new data ~\cite{feng2021survey}. Data augmentation aims to address practical data challenges related to model training. It is applicable to scenarios such as training with low-resource languages  ~\cite{xia2019generalized}, rectifying class imbalance ~\cite{chawla2002smote}, mitigating gender bias ~\cite{zhao2018gender}, and few-shot learning ~\cite{wei2021few,dai2023chataug}. 

Some data augmentation methods incorporate knowledge infusion. For example, Feng et al. ~\cite{feng2020genaug} used WordNet ~\cite{miller1995wordnet} as the knowledge base to replace words with synonyms, hyponyms and hypernyms. Another study ~\cite{grundkiewicz2019neural} extracts confusion sets from the Aspell spellchecker to perform synthetic data generation in an effort to enhance the training data, which consists of erroneous sentences used for training a neural grammar correction model. 

However, there is limited research on the efficacy of applying data augmentation to large pre-trained language models ~\cite{feng2021survey}. In fact, some data augmentation methods have been found to have limited benefit for large language models ~\cite{feng2021survey,longpre2020effective}. For example, EDA ~\cite{wei2019eda}, which consists of 4 operations (synonym replacement, random insertion, random swap, and random deletion), provides minimal performance enhancement for BERT ~\cite{devlin2019bert} and RoBERTa.

Nonetheless, researchers ~\cite{feng2021survey} advocate for more work to explore scenarios in which data augmentation is effective for large pre-trained language models because some studies ~\cite{shi2021substructure} demonstrate results contrary to the claims of ~\cite{longpre2020effective}.

Newly released language models, such as ChatGPT and GPT-4, demonstrate impressive language understanding and reasoning abilities, effectively executing tasks such as grammatical corrections, rephrasing, and text generation. Capitalizing on these advancements, Dai et al.~\cite{dai2023chataug} proposed a ChatGPT-based text data augmentation method that generates high-quality augmented data by rephrasing the original text into multiple semantically similar variants with different representations and styles for training the BERT model on a downstream classification task. Experiments conducted on Amazon customer reviews, medical symptom descriptions, and the PubMed 20K~\cite{dernoncourt2017pubmed} datasets have shown significant improvements in text classification. However, as ChatGPT is pre-trained solely on general domain data, it lacks pre-training knowledge in the agriculture domain. Therefore, it may encounter difficulties generating high-quality augmented text data specific to the agriculture domain.

In this study, we investigate the effectiveness of data augmentation with knowledge infusion and apply our method to the answer selection task scenario. We find that our method significantly improves semantic matching performance. 

\subsection{Answer Selection}
Answer selection refers to the task of finding the correct answer among a set of candidate answers for a specific question. For example, given the question "What is the capital of France?", a solution to this task is required to select the correct answer among the following choices:

\begin{itemize}
  \item A) Paris is the capital of France.
  \item B) Paris is the most populous city in France.
  \item C) London and Paris are financial hubs in Europe.
\end{itemize}

In this case, the first answer should be selected. It is clear that matching words or phrases is not sufficient for this task. 

A common approach is to formulate this problem as a ranking problem such that candidate answers are assigned ranking scores based on their relevance to the question. Earlier work primarily relies on feature engineering and linguistic information ~\cite{yih2013question}. However, the advancement of deep learning introduces powerful models ~\cite{ruckle2019coala,laskar2020contextualized} that outperform traditional methods without the need of manual efforts or feature engineering. 

In this study, our goal is to establish valid mappings between food descriptions and nutrition data. We formulate this task as an answer selection problem and demonstrate the superiority of our method over baselines.

\label{sec:rw}


\section{AgriBERT: Training and Methodology}
\subsection{Domain Specific Language Model}
Language models are powerful tools for a variety of NLP applications. Given a particular task in a specific domain, a language model becomes more effective if it is trained on corpora that contain a large amount of text in the same domain. 
Such practices have already been adopted by various domains in the literature, where BioBERT\cite{lee2020biobert} and FinBERT \cite{huang2022finbert} are successful examples of training domain-specific language models in biomedical and financial domains respectively. 
Motivated by the fact that there is a lack of corpora or pre-trained models in the agricultural domain, in order to produce vocabulary and word embeddings better suited for this domain than the original BERT, we build upon previous research and collect 46,446 articles related to food and agriculture which contains more than 300 million tokens. We then use this corpus to train a BERT model from scratch (more details about the dataset are provided in Section~\ref{subsub:data}). We train our model using the standard procedure of masked language modeling, which involves masking a certain fraction of words in a sentence and requiring the model to predict the masked words based on other words in the sentence. This approach enables the model to learn meaningful sentence representations and can be used for various NLP applications in the agricultural field.

\begin{figure}[t]
\begin{center}
\centerline{\includegraphics[width=\columnwidth]{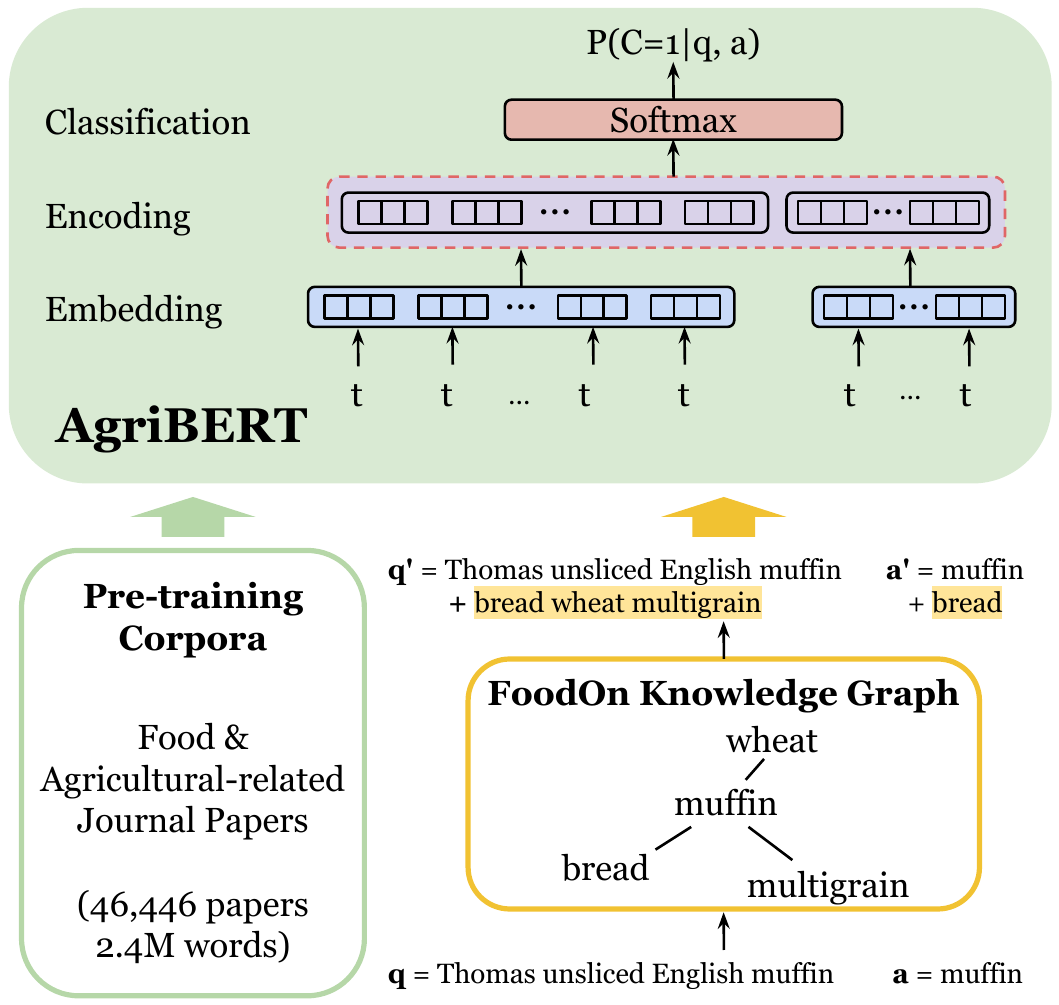}}
\caption{The overall framework of AgriBERT which is trained on agriculture literature from scratch. AgriBERT is evaluated on the answer selection task. The answer selection component has two inputs: a question and an answer, and before we input them into the framework, we add new entities to them from an external source of knowledge such as Wikidata or FoodOn. The output of the framework is a score (a probability) which is used for ranking the answers.}
\label{framework}
\end{center}
\end{figure}

\subsection{Answer Selection Problem Definition}
The evaluation of trained language models requires downstream tasks in the domain. For instance, most biomedical language models are evaluated on named entity recognition tasks with medical datasets. 
Semantic matching, as a technique to determine whether two sentences have similar meanings, has important practical values in various domains, e.g., agriculture. Due to the lack of benchmark NLP datasets in the agriculture domain, we construct our own agriculture benchmark dataset and evaluate our model on the semantic matching task. 
More specifically, in this work, we formulate this task as an answer selection problem, where we assign all USDA descriptions (answers) to each Nielsen product description (question) and select $R$ random incorrect USDA description per product description as negative samples where $R << D$, and $D$ is the total number of USDA descriptions. 
Table \ref{tab:example} provides an example of how we extend the dataset when $R=2$. Let $q$ denote a question and $a$ denote an answer. We can formulate the answer selection task as follows:
\begin{equation}
\label{eq:agribert}
S(q, a) = \text{AgriBERT}(q, a)    ,
\end{equation}
where the function $S$ outputs a score (a probability) for each pair of question and answer. Then, the selected answer $a^*$ for a question $q$ is:
\begin{equation}
a^* = \arg\max_{a \in R} S(q, a)  .
\end{equation}

\subsection{Knowledge Infused Finetuning}
As discussed in Section \ref{sec:rw}, prior research has demonstrated successful integration of relevant information from external knowledge sources to enhance the performance of downstream tasks, such as incorporating facts (i.e., a curated triple extracted from a knowledge graph in the form of \texttt{(entity,relation,entity)}) from knowledge graphs~\cite{liu2020k}, or injecting refined entities extracted from text to a knowledge graph~\cite{rezayi2021edge}. In our setting, where answer selection is the task at hand and the training dataset is limited in size, we propose augmenting both the questions and answers with external knowledge to improve the performance of the answer selection module. 

Discovering relevant external knowledge can be accomplished via different mechanisms, such as entity linking~\cite{sun2018open,gritta2018melbourne,xiong2020pretrained,yasunaga2022deep}, querying~\cite{liang2017neural}, calculating similarity~\cite{karpukhin2020dense}, etc. In this paper, we propose utilizing entity linking and querying for obtaining pertinent external knowledge. In entity linking, the goal is to identify and associate all the named entities in a text with corresponding entities in a knowledge graph. There exist many mature entity linking tools such as Stanford-UBC Entity Linking tool \cite{chang2010stanford}, DBpedia Spotlight \cite{mendes2011dbpediaspotlight}, TAGME \cite{piccinno2014tagme}, etc. This practice is proved to be an effective solution in our scenario. However, one drawback of this approach is that the entity recognition and entity linking algorithms are typically trained on general text corpora like Wikipedia, which may not align well with our specific requirements. So it is essential to reconfigure these tools to suit our purposes rather than relying on them as they are. Hence, we consider a domain-specific knowledge graph, e.g., FoodOn~\cite{dooley2018foodon}, and obtain new knowledge by querying it using the keywords in the text of question-answer pairs. We simply append new entities to the end of questions or answers. More details about this knowledge graph will be provided in Section \ref{subsub:data}. Figure \ref{framework} illustrates our proposed framework. We can modify Equation \ref{eq:agribert} as follows to account for the new entities as follows:
\begin{equation}
\label{eq:enhanced_agribert}
S(q, a) = \text{AgriBERT}(q\mathbin\Vert E_q, a\mathbin\Vert E_a)    ,
\end{equation}
where $E_{q}$ and $E_{a}$ represent the entities enriched from external sources such as Wikidata or FoodOn for the question and answer, respectively, and $\mathbin\Vert$ is the concatenation operator. Please refer to Algorithm \ref{alg:answer_selection} for the detailed procedure of mapping retail scanner data to USDA descriptions using AgriBERT and external knowledge.

\begin{algorithm}[t]
\caption{Knowledge Infused Answer Selection}
\label{alg:answer_selection}
\begin{algorithmic}[1]
\Require Nielsen product descriptions $Q$, 
\Require USDA descriptions $A$, 
\Require number of incorrect descriptions per product $S$, 
\Require pre-trained AgriBERT model, 
\Require external knowledge sources.

\For{each product description $q \in Q$}
    \State Select $R$ samples from $A$ as negative samples
    \For{each random USDA description $a \in R$}
        \State $E_{q} \gets$ \text{obtain entities from external sources for $q$}
        \State $E_{a} \gets$ \text{obtain entities from external sources for $a$}
        \State $S(q, a) \gets$ \text{AgriBERT}($q\mathbin\Vert E_{q}, a\mathbin\Vert E_{a}$)
    \EndFor
    \State $a^* \gets \arg\max_{a \in A} S(q, a)$ 
\EndFor

\end{algorithmic}
\end{algorithm}

\label{sec:method}

\section{Enhancing AgriBERT}
We introduce an innovative approach aimed at further improving the performance of our AgriBERT model. Recognizing that effective language model training relies heavily on the quality and diversity of training data, we turn to Generative Pre-trained Transformers (GPT)~\cite{radford2018improving} to augment our dataset. Specifically, we engineer creative and diverse prompts to guide GPT in generating text samples that closely align with the agricultural domain. This process not only enhances the variety of our training data but also provides AgriBERT with more nuanced and domain-specific contexts to learn from. Through this data augmentation technique, we aim to make AgriBERT more robust and versatile, improving its ability to generalize 
across various agricultural text processing tasks.

To enhance the quality and diversity of our training data, we utilize ChatGPT (gpt-3.5-turbo) \cite{openai2022chatgpt} to guide the enrichment process of our dataset through two carefully designed prompts. The first prompt instructs ChatGPT to ``Expand the semantic space of the query $q$ by generating $d$ related words.'' The aim here is to explore and expand the semantic field around a given query to include a broader range of related concepts and keywords. This strategy not only enhances the diversity of our data, but also enriches the context around each query, enabling AgriBERT to better learn and understand the agricultural terminology. The generation of related words effectively provides a more expansive representation of the query.

The second prompt instructs ChatGPT to ``Rephrase $q$''. This approach provides a valuable method for data augmentation, as it encourages linguistic diversity and challenges the model to understand the same query expressed in various ways. It can help the model to become more resilient to a range of phrasings and wordings, thus improving its robustness and versatility. By learning to understand the same question posed in multiple different ways, AgriBERT becomes more adept at generalizing its knowledge to new, unseen queries, which is a crucial aspect of performance in various agricultural text processing tasks. Together, these two prompts provide an effective method for diversifying and enriching our training dataset, and we hypothesize that their application will result in improved performance of our AgriBERT model. Table \ref{tab:prompts} illustrates several examples to show how these two prompts are used to expand the semantic space and rephrase a given input sentence.

\begin{table*}[t]
\centering
\caption{Sample inputs and their outputs for both prompts in data augmentation}
\label{tab:prompts}
\begin{tabularx}{\textwidth}{@{}XXX@{}}
\toprule
\textbf{Input} & \textbf{Prompt 1:} Expand the semantic space & \textbf{Prompt 2:} Rephrase \\ \midrule
nestle 100 grand milk choc caramel bar caramel 1.5oz & sweet, chocolate, candy & The Nestle 100 Grand Milk Chocolate Caramel Bar measures 1.5oz and contains caramel. \\ \midrule
three musketeers milk chocolat nougat plastic bag iwp 6ct 12.78oz & confectionery, packaging, snack & A 6-count package of 12.78oz milk chocolate nougat bars from Three Musketeers, contained in a plastic bag. \\ \midrule
twisted tea hard iced tea malt beverage malt beverage 5 percent long neck btl in box 144oz & brewing, alcohol, carbonation & This box contains a long-necked bottle of Twisted Tea, which is a 5\% alcohol malt beverage with a total volume of 144 ounces. \\ \bottomrule
\end{tabularx}
\end{table*}

\label{sec:augment}

\section{Experiment}
\begin{table}[t]
    \centering
    \caption{Basic statistics of our dataset compared with two benchmark datasets in the standard language modeling field.}
    \label{tab:stat}
    \resizebox{\columnwidth}{!}{
    \begin{tabular}{lllll}\toprule
        Dataset & Articles & Tokens & Words & Size \\ \midrule
        Penn Treebank & - & 887,521 & 10,000 & 10MB \\
        WikiText-103 & 28,475 & 103,227,021 & 267,735 & 0.5GB \\
        Agriculture corpus & 46,446 & 311,101,592 & 2,394,343 & 4.0GB \\ \bottomrule
    \end{tabular}}

\end{table}

\subsection{Datasets}
We employ several datasets for training the language models, evaluating the trained language model, and augmenting the downstream dataset. In this section, we briefly introduce these datasets. 

\subsubsection{Language Model Pre-Training Datasets}
\label{subsub:data}
Our main dataset is a collection of 46,446 food- and agricultural-related journal papers. We downloaded published articles from 26 journals and converted the pdf files to text format to be used in the masked language modeling task. We also cleaned the dataset by removing URLs, emails, references, and non-ASCII characters. In order to compare the contributions of different components of our model, we consider a secondary dataset for training (WikiText-103) that contains articles extracted from Wikipedia. This dataset also retains numbers, case, and punctuation, which is similar to our dataset described above. Statistics of the two datasets are provided in Table \ref{tab:stat}. We also include Penn Treebank dataset~\cite{marcus1994penn}, which is another common dataset for the task of language modeling, as a reference. 

\subsubsection{Answer Selection Dataset} 
We use two different data sources for this part. First, we use the consumer panel product from the Nielsen Homescan data. Nielsen provides very granular data on the food purchases from the stores at the product barcode or Universal Product Code (UPC) with detailed attributes for each UPC, including UPC description. While scanner data come with some nutrition-related product attribute variables, this information is not sufficient to examine the nutritional quality. To address this issue, we link product-level data from Nielsen with the USDA Food Acquisition and Purchase Survey (FoodAPS) which supplements scanner data with detailed nutritional information. The survey contains detailed information about the food purchased or otherwise acquired for consumption during a seven-day period by a nationally representative sample of 4826 US households. The FoodAPS matched 32000+ barcodes with the Food and Nutrient Database for Dietary Studies (FNDDS) food codes of high quality. The linked data set has UPC descriptions for each product and the corresponding FNDDS food code. In addition, the final data set has full information needed to construct diet quality indexes to evaluate the healthfulness of overall purchases.

\subsubsection{External Source of Knowledge}
We use the FoodOn knowledge graph for the question-answer augmentation purposes. FoodOn is formatted in the form of Web Ontology Language (OWL) \cite{antoniou2004semantic}. The OWL ontology provides a globally unique identifier (URI) for each concept which is used for lookup services and facilitates the query processing system. Most of FoodOn’s core vocabulary comes from transforming LanguaL, a mature and popular food indexing thesaurus~\cite{dooley2018foodon}. That is why FoodOn is a unique and valuable resource for enhancing our language model. 

\subsection{Metrics}
Since the output of the evaluation task on the answer selection dataset is a ranked list of answers per question, we require metrics that take into account the order of results. That is why we propose to use precision@1 (P@1) and Mean Average Precision (MAP).

\noindent\textbf{P@1} answers the following question: ``Is the top-ranked item in the returned list a relevant item?'' If the top-ranked item is relevant, $P@1$ is 1 and if it is not relevant, $P@1$ is 0. TO caluclate $P@1$, we sort the selected answers based on the final similarity scores and we count how many times the top answer is correctly selected. 
$$P@1 = \sum_{i = 1}^{|N|} 1 \, \text{if } rank_{a_i} ==1$$
where N is the set of all questions. Note that, while this metric provides valuable insights into the performance of the ranking system at the top of the list, it does not consider the relevance of lower-ranked items. Thus we also calculate and report Mean Average Precision (MAP).

\noindent\textbf{MAP} measures the percentage of relevant selected answers, it takes into account the precision at every rank in the list where a relevant item is found. This is in contrast to $P@1$, which only considers the precision of the top-ranked item. By doing so, MAP values the ranking system's ability to rank relevant items higher. Given a ranked list of selected answers per question we mark them as relevant if they are correctly selected and calculate $\text{AP}$ as follows:
$$\text{AP}=\frac{1}{n}{\sum_{i=1}^n (P(i)\times\text{rel}(k))},$$
where n is the set of all selected answers, $\text{rel}(k)\in\{0,1\}$ indicates if the answer is relevant or not, and $P(i)$ is the precision at $i$ in the ranked list. Once we obtain $\text{AP}$ for each question we can  average across all questions to find $\text{MAP}$:
$$\text{MAP}=\frac{1}{|N|}{\sum_{q=1}^{|N|}\text{AP(q)}},$$
where $N$ is the set of all questions.

\subsection{Baselines}
To study the impact of different language model pre-training datasets and strategies, different answer-select datasets, we conduct a series of ablation studies. 
We consider the following scenarios:

\begin{itemize}
  \item \textbf{kNN:} We compute the embeddings\footnote{We use sentence-transformer library for this task: \url{https://github.com/UKPLab/sentence-transformers}} of the Nielsen product descriptions and USDA descriptions. For each vector belonging to the product description embedding space, we find the most similar vector from the USDA description embedding space. This naive approach is effective if the number of unique USDA descriptions is small. However, this does not hold in our case (Row 1).
  \item \textbf{BERT\textsuperscript{p}:} We use the pre-trained BERT model without any modification (Row 2).
  \item \textbf{BERT\textsuperscript{p}}: We further finetune the pre-trained BERT using WikiText-103 (Row 3) and Agricultural Corpus (Row 6).
  \item \textbf{BERT\textsuperscript{s} on WikiText-103:} We train the BERT model from scratch using WikiText-103 with additional enhancements by connecting the existing entities to Wikidata (Row 4) and FoodOn knowledge graph (Rows 5).
  \item \textbf{BERT\textsuperscript{s} on Agricultural Corpus:} We train the pre-trained BERT model from scratch using our Agriculture corpus (Rows 7).
  \item 
  \item \textbf{BERT\textsuperscript{s} + Entity Linking:} We train the BERT model from scratch using Agricultural Corpus with additional enhancement by connecting the existing entities to Wikidata (Row 8), and Food on considering different numbers of entities (Rows 9-11).
  \item \textbf{BERT\textsuperscript{s} + ChatGPT:} We train the BERT model from scratch using Agricultural Corpus with additional enhancement by manually defined prompts p1 and p2 (Rows 12-13).
\end{itemize}



\begin{table*}[t]
    \centering
        \caption{Examples to demonstrate the quality of added entities to the text of product descriptions. For Wikidata we use entity linking and we present here the top linked entity (highest confidence score). For FoodOn we use SPARQL to query the ontology and the first outcome is listed here.}
    \label{tab:qual}
    \resizebox{.7\textwidth}{!}{
    \begin{tabular}{@{}lll@{}}\toprule
        Sentence & Wikidata entity & FoodOn entity \\ \midrule
        nestle nido powder infant formula & nestle & rice powder \\
        aunt jemima frozen french toast breakfast entree  & aunt jemima & frozen dairy dessert \\
        woodys hickory barbecue cooking sauce & woody's chicago style & hickory nut  \\
        sour punch sour watermelon fruit chew straw & sour punch & sour milk beverage \\
        philly steak frozen beef sandwich steak &  philly steaks & wagyu steak \\ 
        yoplait original rfg harvest peach yogurt low fat & yoplait & creamy salad dressing \\ \bottomrule
    \end{tabular}}

\end{table*}

\subsection{Experimental Settings}
Once the extended dataset is generated, we can apply any answer selection method to the dataset. There are a number of studies in the literature on this topic, including COALA~\cite{ruckle2019coala}, CETE~\cite{laskar2020contextualized}, MTQA~\cite{deng2019multi}, and many more, among which CETE is considered state-of-the-art in the answer selection task by the ACL community\footnote{Reported here: \url{https://aclweb.org/aclwiki/Question_Answering_(State_of_the_art)}}. CETE implements a transformer-based encoder (e.g., BERT) to encode the question and answer pair into a single vector and calculates the probability that a pair of question/answer should match or not\footnote{The code for this study is open source and available for public use: \url{https://github.com/tahmedge/CETE-LREC}}. 

For the entity linking process, we use the implementation proposed by Wu et al. ~\cite{wu2020scalable}, called BLINK. BLINK is an entity linking Python library that uses Wikipedia as the target knowledge base. Moreover, in order to send efficient SPARQL queries to the FoodOn knowledge graph, 
we use ROBOT, 
a tool for working with Open Biomedical Ontologies\footnote{Find it here: \url{https://github.com/ontodev/robot}}. Additionally, since we aim at simulating a setting where the number of labeled training samples is small, we use 20\% of the dataset for training and the remaining 80\% for the test. We believe this is a more realistic scenario in real-world applications. 

\begin{table}[t]
    \centering
        \caption{Test performances of all models trained on different datasets for the task of answer selection. for the kNN model we use sentence-transformers to compute embeddings.  EL stands for Entity Linking and bold numbers indicate the best performance. BERT\textsuperscript{p} indicates a pre-trained BERT and BERT\textsuperscript{s} means training a BERT model from scratch. $n$ indicates the number of external entities we include in the text of questions and answer when using FoodOn KG. p1 and p2 indicate which prompt we use to instruct ChatGPT to enrich the training data.}
    \label{tab:result}
    \resizebox{\columnwidth}{!}{
    \begin{tabular}{@{}l|llcc@{}}\toprule
        \# &Training Dataset & Model & MAP & P@1 \\ \midrule
        1 & - & kNN & 26.70 & 14.49 \\
        2 & - & BERT\textsuperscript{p} & 27,77 & 10.88 \\ \midrule
        3 & WikiText-103 & BERT\textsuperscript{p} & 28.03 & 11.12 \\
        4 & WikiText-103 & BERT\textsuperscript{s}+EL (Wikidata) & 27.36 & 10.09 \\
        5 & WikiText-103 & BERT\textsuperscript{s}+FoodOn (n=1) & 28.78 & 24.83 \\ \midrule
        6 & Agricultural Corpus & BERT\textsuperscript{p} & 29.72 & 12.71 \\
        7 & Agricultural Corpus & BERT\textsuperscript{s} & \textbf{44.21} & 22.72 \\
        8 & Agricultural Corpus & BERT\textsuperscript{s}+EL (Wikidata) & 42.33 & 21.52 \\
        9 & Agricultural Corpus & BERT\textsuperscript{s}+FoodOn (n=1) & 31.54 & 47.89 \\
        10 & Agricultural Corpus & BERT\textsuperscript{s}+FoodOn (n=3) & 30.65 & 49.80 \\
        11 & Agricultural Corpus & BERT\textsuperscript{s}+FoodOn (n=5) & 29.91 & \textbf{49.98} \\
        12 & Agricultural Corpus & BERT\textsuperscript{s}+ChatGPT (p1) & 32.03 & 48.78 \\
        13 & Agricultural Corpus & BERT\textsuperscript{s}+ChatGPT (p2) & 30.19 & 46.91 \\ \bottomrule
    \end{tabular}}
    \vspace{-2mm}
\end{table}

\subsection{Results}
As Table \ref{tab:result} presents, not surprisingly, the best performance is obtained when the language model is trained on the agricultural corpus. We summarize the main observations as follows: In terms of Mean Average Precision (MAP), the BERT model trained from scratch (BERT\textsuperscript{s}) on the Agricultural Corpus outperforms all other models, achieving the highest MAP of 44.21 (entry 7). This suggests that the BERT model significantly benefits from domain-specific training data. It is also notable that BERT\textsuperscript{s} with Entity Linking (EL) using Wikidata performed nearly as well with a MAP of 42.33 (entry 8). The least-performing model in terms of MAP is the kNN model with a MAP of 26.70 (entry 1).

Regarding the Precision at rank 1 (P@1), the BERT model trained from scratch with data augmentation using FoodOn (with n=5) and ChatGPT prompt 1 (p1) show high performance with P@1 scores of 49.98 (entry 11) and 48.78 (entry 12), respectively. This shows that enhancing the training data with additional semantic information from FoodOn and GPT-generated prompts improves the model's ability to select the correct answer at the top rank. Interestingly, even though the BERT model trained from scratch on the Agricultural Corpus achieved the highest MAP, it did not achieve the highest P@1. This suggests that while this model is good at ranking relevant answers highly across all retrieved answers, it may not always select the correct answer as the top-ranked answer.


By comparing rows 5 and 9 we can see that training on the Agricultural Corpus led to a significant improvement in both the MAP and P@1 scores. This suggests that the Agricultural Corpus is a more suitable training dataset for this task compared to WikiText-103.

We also investigate the number of external entities that we include in the text of questions and answers as shown by entries 9, 10, and 11 in Table \ref{tab:result}. 
We can see that the growth of the number of external entities $n$ will lead to a decrease in the MAP score and an increment in the P@1. This suggests that incorporating related entities from a relevant knowledge source helps to find the correct match in 50\% of the times but sometimes it misleads the answer selection module and ranks the correct match lower which means it decreases the MAP score. Table \ref{tab:qual} provides some examples of augmented sentences by Wikidata and FoodOn knowledge sources. As this table presents, linking the food description to Wikidata entities can easily go wrong. 
For instance, in the first three rows, the food descriptions are linked to brand names\footnote{\url{https://en.wikipedia.org/wiki/Aunt_Jemima}}. In contrast, this does not happen when we query the FoodOn KG, as the entities are purely food related. Additionally, FoodOn entities contain food-related adjectives such as frozen, creamy, etc. which help in matching the food descriptions to nutrition data.

Additionally, data augmentation using either FoodOn (a domain-specific knowledge graph) or ChatGPT (a powerful language model) yields comparable results. This resemblance reinforces the recent discourse highlighting the conceptual similarities between knowledge graphs and language models. Knowledge graphs, like FoodOn, provide structured and explicit knowledge which can be selectively used to augment the data, aiding in particular tasks. On the other hand, language models, such as ChatGPT, implicitly capture a vast amount of knowledge from the data they are trained on, effectively serving as a form of knowledge base themselves.

This connection is further explored 
by Petroni et al. ~\cite{petroni2019language}, which investigates the feasibility of language models as a source of knowledge. The authors suggest that the knowledge present in the parameters of a language model can be effectively queried, much like a traditional knowledge graph, underlining the potential of language models to serve as knowledge bases. In our case, the performance similarity of FoodOn and ChatGPT data augmentation techniques indicates that both explicit, structured knowledge and implicit, context-rich knowledge can contribute to the enhancement of domain-specific language model performance. This suggests that combining structured knowledge graphs with richly trained language models could unlock further advancements in model performance and capability.

\subsection{GPT-based Baseline}
In this subsection, we present an exploratory investigation that focuses on the task of answer selection utilizing the capabilities of the GPT-3.5 language model. Given a query and a set of candidate documents, the objective of the task is to identify the document that is most relevant to the query. For the purpose of this study, we have chosen a subset of 100 queries to analyze the performance of GPT-3.5 in this context. It is important to note that the scale of evaluation is limited when compared to other baselines, such as AgriBERT language model, which was tested against the entire dataset. While our analysis may not be as comprehensive as those conducted with other models, it offers valuable insights into the model's potential for this task. Our intention is not to evaluate the efficacy of GPT-3.5, but rather to o highlight the potential of GPT-3.5 in scenarios where it could be deployed independently.

To facilitate this investigation, we design a specific prompt aimed at guiding the GPT-3.5 model's response. The prompt first presents a 'Given' section, which includes a specific query and a list of documents. The `Query' in this context is a product description. The `Documents' section lists ten potential options, from USDA descriptions (see Table \ref{tab:example}). The `Task' section provides a clear instruction to the model, asking it to ``Rank the documents in order of relevance to the given query''. This simple, direct instruction enables the model to understand the requested operation - that is, to assess each document in terms of its relevance to the query and provide a ranking based on this assessment. The structure of the prompt is as follows:
\begin{verbatim}
Given:
Query: domino white sugar granulated 1lb

Documents:
1: salsa, red, commercially-prepared
2: cookie-crisp 
.
.
.
10: sugar, white, granulated or lump

Task:
Rank the documents in order of relevance 
to the given query (no explanation required).

\end{verbatim}

Here is a sample output:
\begin{verbatim}
9, 1, 10, 2, 6, 4, 7, 8, 5, 3
\end{verbatim}

We can easily parse the response and obtain MAP and P@1 for 1,000 randomly selected queries as follows: MAP=73.33, P@1=60.12. GPT-3.5, even without being fine-tuned for the agricultural domain, outperforms the AgriBERT model, suggesting the strength of large, general-purpose language models in the answer selection task. The higher MAP score of 0.73 for GPT-3.5 compared to AgriBERT's 29.91 suggests that GPT-3.5 is consistently ranking the correct document higher in the list across all queries. This could potentially be attributed to the larger size and diverse training data of GPT-3.5. On the other hand, the smaller difference in P@1 scores suggests that both models are fairly competitive when it comes to ranking the most relevant document at the top. This could mean that both models are capable of identifying the most relevant document, but GPT-3.5 may be better at providing a more accurate overall ranking.

\subsection{Experiment: Cuisine Prediction}

In this section, we investigate the performance of AgriBERT in the task of cuisine prediction. The experimental results are based on a unique dataset provided by Yummly\footnote{\url{https://www.yummly.com/}}, which was originally featured in a Kaggle competition\footnote{\url{https://www.kaggle.com/competitions/whats-cooking/overview/description}}. The task aims at classifying recipes into different cuisines such as Greek, Indian, Italian, and more, using their ingredient lists. Essentially, we treat this task as a sentence classification task in which we use different sentence representation learning methods to generate feature representations of different recipes and then feed these representations into a classifier to classify the given recipes into different cuisines.

However, it is important to note that we were unable to use the Kaggle test set for evaluation purposes due to the unavailability of label information. Therefore, we split the provided training set, reserving 20\% of it as our test set. This approach means that our results are not directly comparable with the Kaggle leaderboard since we used a different test set. In the experiment, we leverage the domain-specific knowledge and contextual understanding of agricultural text offered by AgriBERT to predict the cuisine category of recipes. We then compare its performance with other feature representations and classifiers commonly used in natural language processing.

\begin{table*}[t]
\centering
\label{tab:cc}
\caption{Performance comparison of various feature representations and classifiers in the cuisine prediction task. The values in the table represent the F1 scores achieved by each combination. It is observed that AgriBERT consistently outperforms other models, demonstrating its effectiveness in this task.}
\begin{tabular}{@{}lccccc@{}}
\toprule
Feature                   & Logistic Regression & SVM              & Decision Tree    & Random Forest    & MLP              \\ \midrule
TF-IDF                    & 57.67\%             & 47.64\%          & 28.25\%          & 41.93\%          & 70.76\%          \\
agri-sentence-transformer & 64.34\%             & 65.87\%          & 27.71\%          & 42.55\%          & 61.37\%          \\
all-mpnet-base-v2         & 66.29\%             & 71.10\%          & 43.00\%          & 57.77\%          & 73.51\%          \\
AgriBERT                 & \textbf{77.52\%}    & \textbf{79.63\%} & \textbf{61.99\%} & \textbf{73.18\%} & \textbf{75.79\%} \\ \bottomrule
\end{tabular}
\end{table*}

Table \ref{tab:cc} compares the performances of 
different combinations of feature representations and classifiers. 
We use four feature representations: 
\begin{enumerate}
    \item \textbf{TF-IDF}: a traditional and simple approach for text vectorization based on term frequency and inverse document frequency.
    \item \textbf{agri-sentence-transformer}: agri-sentence-transformer is a BERT-based language model further pre-trained from the checkpoint of SciBERT. This model was trained on a balanced dataset composed of both scientific and general works in the agriculture domain, encompassing knowledge from different areas of agriculture research and practical knowledge. The corpus contains 1.2 million paragraphs from the National Agricultural Library (NAL) in the US, and 5.3 million paragraphs from books and common literature from the Agriculture Domain.
    \item \textbf{all-mpnet-base-v2}: a variant of the MPNet architecture, a transformer-based model developed by Microsoft Research. The `all-MPNet-base-v2` configuration includes 12 layers, with a hidden size of 768, and 12 attention heads.
    \item \textbf{AgriBERT}: the version of AgriBERT that trains a BERT architecture from scratch (Row 7).
\end{enumerate}

We evaluate the performance of these feature representations on five different classifiers such as logistic regression, support vector machine (SVM), decision tree, random forest, and multilayer perceptron (MLP). 
Although the choice of classifier is not our focus, it's interesting to see how different feature representations perform with the same classifier.

To begin, the TF-IDF feature representation, a traditional and simple approach for text vectorization, appears to perform reasonably well with the MLP Classifier, yielding an F1 score of 70.76\%. However, its performance seems to drop significantly with the other classifiers, suggesting that it may not be the most robust choice of feature representations across various classification methods.

In terms of agri-sentence-transformer, 
despite this domain-specific knowledge, the model performs best with the Support Vector Machine (SVM) classifier, achieving an F1 score of 65.87\%. Interestingly, it does not outperform the general-purpose PLM, all-mpnet-base-v2, in the task of cuisine prediction. 

Finally, the most effective among all feature representations is AgriBERT. Regardless of the choice of classifier, AgriBERT consistently exhibits superior performance compared to the other models, achieving its highest F1 score of 79.63\% with the SVM classifier. This implies that AgriBERT is notably adept at identifying the essential features needed for cuisine prediction, thereby substantiating the effectiveness of its design and training approach.

Building on the demonstrated effectiveness of our AgriBERT model in both cuisine prediction and semantic matching tasks, we propose that AgriBERT is well-suited for a wider range of applications within the food and agriculture domains. Its ability to capture and understand nuanced relationships within agricultural text positions it as a robust tool for various tasks. These could encompass areas such as ingredient substitution, recipe recommendation, food-related sentiment analysis, and more. The strong performance of AgriBERT on diverse tasks indicates its generalizability, making it a promising resource for future research and application development in the food and agriculture sectors.
\label{sec:exp}

\section{Conclusion}
In this paper, we present a language model called AgriBERT that can facilitate the NLP tasks in the food and agricultural domain. AgriBERT is a BERT model trained from scratch with a large corpus of academic journals in the agriculture field. 
We evaluate our language model on two tasks: semantic matching and cuisine prediction. 
The semantic matching 
aims at matching two databases of food description (Nielsen database and USDA database). We reformulate the problem as an answer selection task and used our language model as a backbone of a generic answer selection module to find the best match. Before feeding the pairs of questions and answers to the model we augmented them with external entities obtained from the FoodOn knowledge graph, a domain-specific ontology in the field of food. We showed that the inclusion of external knowledge can help boost the performance in terms of the more strict P@1 measure but it lowers the performance in terms of mean average precision. 
The cuisine prediction is a task to classify recipes into different cuisines. We show that compare with various existing language representation learning methods including the state-of-the-art all-mpnet-base-v2 model, our AgriBERT can achieve the best performance with different classifiers. 
As a future direction, we plan to investigate more sophisticated approaches for incorporating external knowledge such as refining the knowledge before including it in the text. 
Another future research direction is to fine-tune the existing large language foundation models such as InstructGPT \cite{ouyang2022instructgpt}, ChatGPT, StableLM\footnote{\url{https://github.com/Stability-AI/StableLM}}, and so on on our collected agriculture corpus and investigate their performances on the given agriculture tasks \cite{lu2023agi}.

\label{sec:con}

\bibliographystyle{IEEEtran}
\bibliography{main}

\end{document}